\documentclass[10pt]{article} %
\usepackage[preprint]{tmlr}

\usepackage{hyperref}
\usepackage{url}
\usepackage{epsfig}
\usepackage{graphicx}
\usepackage{amsmath}
\usepackage{amssymb}
\usepackage{graphicx}
\usepackage{multirow}
\usepackage{amsmath}
\usepackage{amssymb}
\usepackage{caption, subcaption}
\usepackage[para]{threeparttable}
\usepackage{url}            %
\usepackage{booktabs}       %
\usepackage{amsfonts}       %
\usepackage{nicefrac}       %
\usepackage{microtype}      %
\usepackage{pifont}
\usepackage{placeins}
\usepackage{xcolor}
\usepackage{gensymb} 
\usepackage{xcolor,colortbl}
\usepackage[capitalize]{cleveref}

\newcommand{\xmark}{\ding{55}}

\title{Adapting Contrastive Language-Image Pretrained (CLIP) Models for Out-of-Distribution Detection}

\author{\name Nikolas Adaloglou  {\tt\small \email adaloglo@hhu.he} \\
\addr Heinrich Heine University, Duesseldorf
\AND 
\name Felix Michels {\tt\small \email felix.michels@hhu.de}  \\
\addr Heinrich Heine University, Duesseldorf
\AND  \name Tim Kaiser  {\tt\small \email tikai103@hhu.de} \\
\addr Heinrich Heine University, Duesseldorf
\AND 
 \name Markus Kollman  {\tt\small \email markus.kollmann@hhu.de}  \\
 \addr Heinrich Heine University, Duesseldorf\\
}

\def\month{MM}  %
\def\year{YYYY} %
\def\openreview{\url{https://openreview.net/forum?id=XXXX}} %

\begin{document}

\crefname{table}{Tab.}{Table}%

\maketitle
\begin{abstract}
We present a comprehensive experimental study on pretrained feature extractors for visual out-of-distribution (OOD) detection, focusing on adapting contrastive language-image pretrained (CLIP) models. Without fine-tuning on the training data, we are able to establish a positive correlation ($R^2\geq0.92$) between in-distribution classification and unsupervised OOD detection for CLIP models in $4$ benchmarks. We further propose a new simple and scalable method called \textit{pseudo-label probing} (PLP) that adapts vision-language models for OOD detection. Given a set of label names of the training set, PLP trains a linear layer using the pseudo-labels derived from the text encoder of CLIP. To test the OOD detection robustness of pretrained models, we develop a novel feature-based adversarial OOD data manipulation approach to create adversarial samples. Intriguingly, we show that (i) PLP outperforms the previous state-of-the-art \citep{ming2022mcm} on all $5$ large-scale benchmarks based on ImageNet, specifically by an average AUROC gain of 3.4\% using the largest CLIP model (ViT-G), (ii) we show that linear probing outperforms fine-tuning by large margins for CLIP architectures (i.e. CLIP ViT-H achieves a mean gain of 7.3\% AUROC on average on all ImageNet-based benchmarks), and (iii) billion-parameter CLIP models still fail at detecting adversarially manipulated OOD images. The code and adversarially created datasets will be made publicly available.
\end{abstract}

\begin{figure}
\begin{center}
\includegraphics[width=1\columnwidth, trim={0 4.1cm 0 0}, clip]{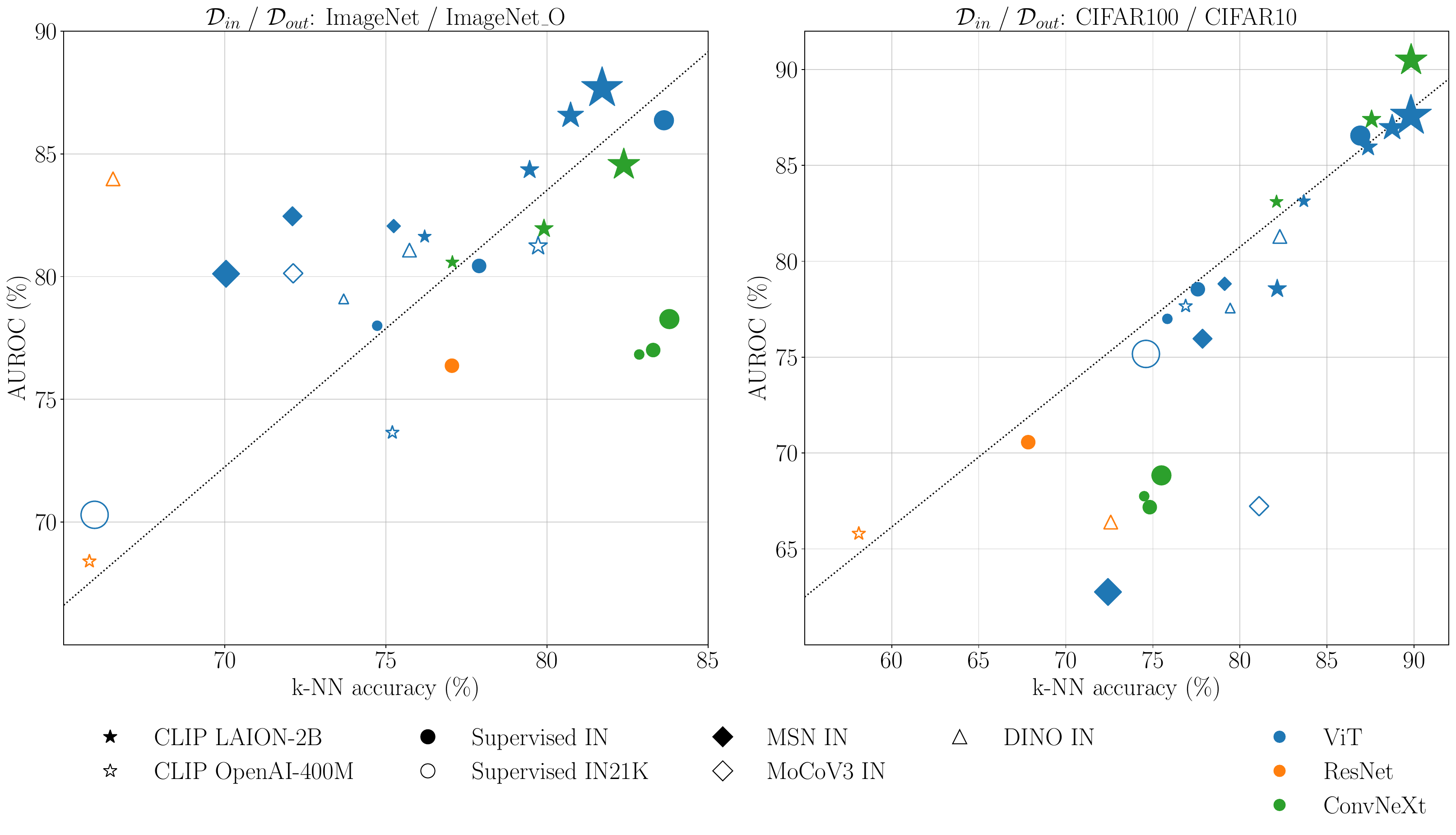}
\includegraphics[width=1\columnwidth]{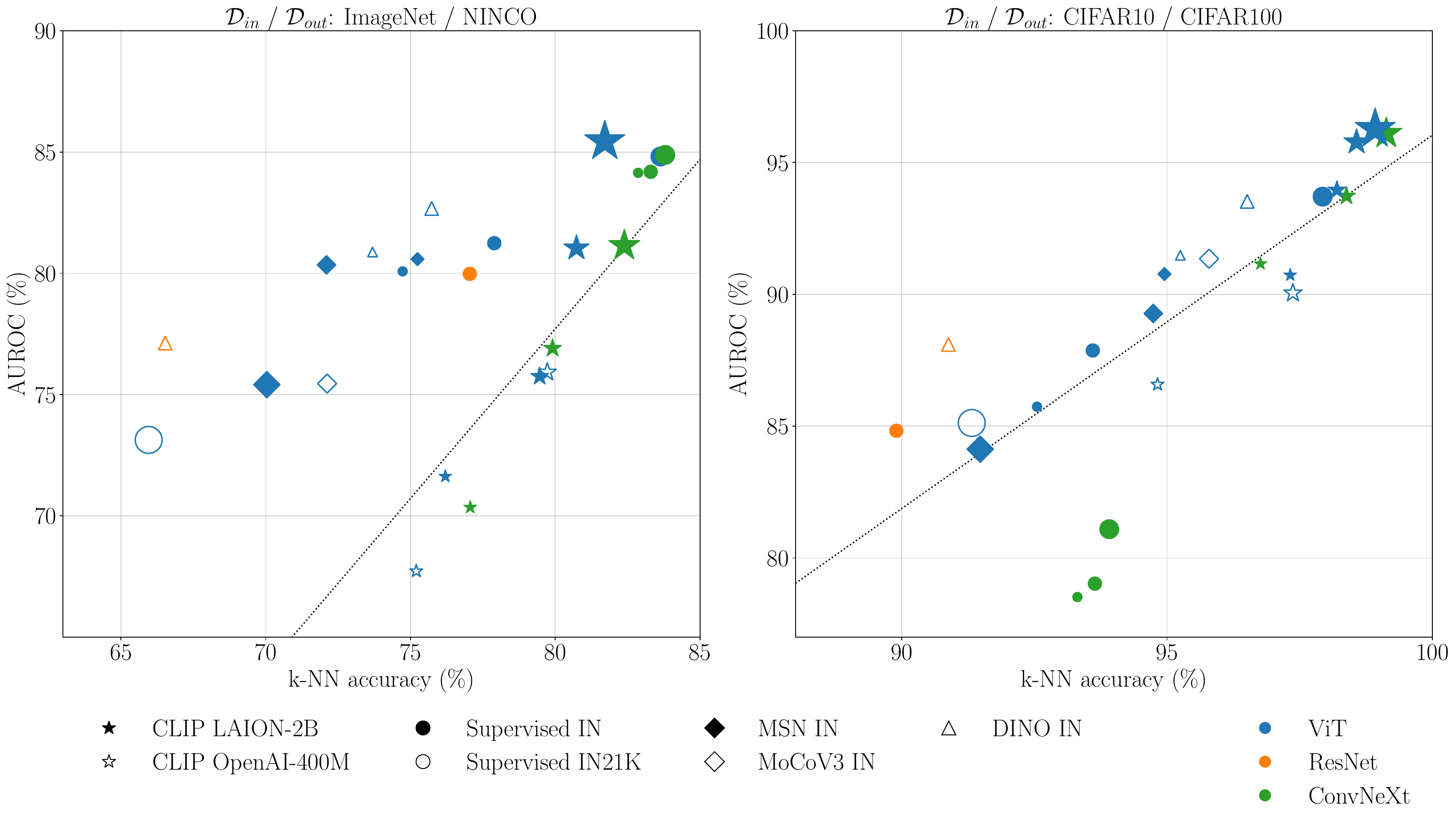}
\end{center}
\caption{\textbf{In-distribution test accuracy using $k=20$ nearest neighbours (k-NN) (x-axis) versus unsupervised out-of-distribution (OOD) detection score (AUROC \%) (y-axis) for ImageNet$\rightarrow$ImageNet-O, CIFAR100$\rightarrow$CIFAR10, ImageNet$\rightarrow$NINCO and CIFAR10$\rightarrow$CIFAR100} \cite{deng2009imagenet, hendrycks2021natural, cifar}. CLIP models \cite{radford2021clip, cherti2022reproducible} exhibit a strong dependence between OOD performance and in-distribution test accuracy even without fine-tuning. The black dotted line is fitted using the CLIP models (starred datapoints), where a coefficient $R^2>0.92$ for all benchmarks was found. The OOD detection score is computed using the top-1 NN cosine similarity. Different colors are utilized for different architectures (ViT \cite{vit}, ConvNeXt \cite{convnext}, ResNet \cite{resnet}) while symbol sizes roughly indicate architecture size (i.e. Small, Base, Large, Huge, Giga). IN indicates ImageNet and IN21K indicates ImageNet-21K \cite{russakovsky2015imagenet}. Best viewed in color.}
\label{fig:diagram}
\end{figure}

\section{Introduction}
Transferring the representations of pretrained vision models has improved the performance on a plethora of image recognition tasks \citep{yosinski2014transferable, tan2018survey, park2023what, adaloglou2023exploring}. To date, these models are trained with various types of supervision, which accelerates training convergence compared to random initialization \citep{he2019rethinking}. Examples include self-supervision \citep{simclr}, natural language supervision \citep{radford2021clip}, weakly-supervised learning \citep{mahajan2018exploring}, or standard supervised learning. Concurrently, \cite{vit} have established vision transformers (ViTs), along with an enormous number of variants \citep{liu2021swin, touvron2021going, beyer2023flexivit}, as a suitable architecture for training large-scale models in the visual domain \citep{dehghani2023scaling}. 

Nevertheless, the applicability of the learned features of pretrained models is crucial and non-trivial, especially for unsupervised downstream tasks \citep{bommasani2021foundational, adaloglou2023exploring}. This work centers on adapting pretrained feature extractors for various visual OOD detection setups, focusing on contrastive language-image pretraining (\emph{CLIP}) models. 

The task of OOD, novelty, or anomaly detection aims at identifying whether a given test sample is drawn from the \textit{in-distribution} (the training set) or an alternative distribution, known as the \textit{out-distribution}. Accurate detection of anomalies is indispensable for real-world applications to ensure safety during deployment \citep{amodei2016concrete,ren2019likelihood}. The detected unfamiliar samples can be processed separately, possibly with a human expert in the loop, rather than making a potentially uncalibrated prediction \citep{guo2017calibration}. Despite significant advances in deep learning, neural networks tend to generate systematic errors for test examples far from the training set \citep{Nguyenieee} or assign higher likelihoods to OOD samples compared to in-distribution samples \citep{nalisnick2018do}.

Recent studies have established a firm connection between the training data distribution accuracy and OOD generalization \citep{hendrycks2021many,wenzel2022assaying, dehghani2023scaling}. A similar connection has been identified for supervised OOD detection when in-distribution training labels are available for training or fine-tuning. Supervised training leads to intermediate representations that likely form tight label-related clusters \citep{fort2021exploring}. An ideal representation for OOD detection should capture semantic properties, such as the pose and shape of an object while remaining sensitive to the properties of the imaging process (e.g. lighting, resolution) \citep{winkens2020contrastive}. 

A suitable choice of visual feature representations is critical for detecting anomalies. However, learning informative representations for unsupervised OOD detection \citep{tack2020csi,sehwag2021ssd,rafiee2022sdns}, where no in-distribution labels are available, is a challenging and active research area \citep{cohen2023out}. Unsupervised methods often adopt self-supervision to learn the in-distribution features by defining pretext tasks such as rotation prediction \citep{gidaris2018rot}. A major milestone in visual representation learning was reached by \cite{simclr} with the development of contrastive and non-contrastive visual self-supervised methods \citep{dino}. Recently, CLIP has enabled learning from vast amounts of raw text \citep{radford2021clip}. Vision-language models offer the unprecedented advantage of zero-shot image classification by leveraging the label names of the downstream dataset. Nonetheless, adapting CLIP models is not straightforward, even in the supervised scenario. For instance, \cite{pham2023combined} showed that fine-tuning CLIP models degrades their robustness against classifying distribution-shifted samples.

Labeled OOD samples are typically unavailable in real-world applications, and the number of in-distribution samples is usually limited. Therefore, external data have been widely employed \citep{rafiee2022sdns} in two ways: a) outlier exposure where the external data is treated as anomalous \citep{hendrycks2018deep}, and b) using models pretrained on auxiliary data \citep{sun2022oodknn}. Outlier exposure leads to performance gains only if the auxiliary data are sufficiently diverse and disjoint from the in-distribution \citep{hendrycks2018deep,liznerski2022exposing}. On the other hand, \cite{hendrycks2020pretrained,hendrycks2021many} showed that pretrained backbones can enhance OOD detection performance and robustness without relying on dataset-specific shortcuts \citep{geirhos2020shortcut}. Consequently, pretrained models are suitable candidates for OOD detection, while \cite{galil2023a} has recently established CLIP models for OOD detection, especially when in-distribution label names are present \citep{ming2022mcm}.

In parallel, several OOD detection methods still rely on similar small-scale benchmarks based on low-resolution images \citep{nvidia2021shifting, rafiee2022sdns, esmaeilpour2022zero}, such as CIFAR \citep{cifar}. \cite{huang2021mos} argued that methods explicitly tuned for these benchmarks may not always translate effectively into larger-scale and real-life applications. Towards this direction, new large-scale and more challenging benchmarks have been introduced \citep{hendrycks2021natural, yang2022openood, bitterwolf2023ninco}, which consider ImageNet \citep{deng2009imagenet} as in-distribution. Finally, even though the robustness against adversarial attacks has been sufficiently explored in image classification \citep{intriguingproperties2014, explainingharnessing, mao2023understanding}, less attention has been given to studying the construction of robust visual OOD detectors \citep{adversarial-ood, adversarialAdam}. Even though several advancements in visual feature extractors have been made and new large-scale OOD detection benchmarks \citep{bitterwolf2023ninco} have been proposed, limited research has been conducted in OOD detection regarding the choice of pretrained model and evaluation scheme, especially regarding CLIP models \citep{ming2022mcm, galil2023a, esmaeilpour2022zero}.

In this paper, we present an experimental study across $25$ feature extractors and several visual OOD detection benchmarks. Using the existing publicly available models, we demonstrate that large-scale CLIP models are robust unsupervised OOD detectors and focus on adapting the representations of CLIP under different OOD detection settings. Under this scope, we examine several OOD detection setups based on the availability of labels or image captions (i.e. in-distribution class names). The core contributions of this work are summarized as follows: 

\begin{itemize}
    \item To investigate whether the dependence between in-distribution accuracy and unsupervised OOD detection performance can be confirmed without in-distribution fine-tuning, we quantify the performance of $25$ pretrained models across $4$ benchmarks (\cref{fig:diagram}). CLIP models exhibit the strongest positive correlation without fine-tuning across all benchmarks ($R^2\text{ coefficient}\geq0.92$). Interestingly, the features of CLIP ViT-G outperform the ones from supervised ImageNet pretraining on ImageNet-based OOD detection. 
    \item To adapt the representations of CLIP for OOD detection, we propose a simple and scalable method called pseudo-label probing (PLP). Here, text-based pseudo-labels are computed using its text encoder based on the maximum image-text feature similarity. We leverage the obtained text-based pseudo-labels to train a linear layer on top of CLIP. PLP surpasses the previous state-of-the-art \citep{ming2022mcm} on $5$ ImageNet benchmarks by an average AUROC gain of 1.8\% and 3.4\% for CLIP ViT-H and CLIP ViT-G, respectively. Moreover, linear probing achieves superior performance on ImageNet-based OOD benchmarks compared to fine-tuning, notably by a mean AUROC gain of 7.3\% using CLIP ViT-H.
    \item Finally, we introduce a novel method that adversarially manipulates OOD images by matching their representations to in-distribution samples and confirm that CLIP ViT-G trained on billions of samples can be easily fooled (86.2\% $\rightarrow$ 50.3\% AUROC deterioration), by changes that are invisible to humans.
\end{itemize}

\section{Related work}
\subsection{Supervised OOD detection methods}
Supervised OOD detection methods rely on the fact that in-distribution classification accuracy is positively correlated with OOD detection performance \citep{fort2021exploring, galil2023a}. For that reason, many OOD detection methods derive anomaly scores from supervised in-distribution classifiers. \cite{hendrycks2016msp} developed the maximum softmax probability (MSP) as OOD detection score, which is frequently used, or its temperature scaled version \citep{liang2018enhancing}. More recently developed scores based on the logits of the in-distribution classifier are the maximum logit \citep{hendrycks2022scalingood} or the negative energy scores by \cite{liu2020energy}. An alternative OOD detection score, which requires the in-distribution labels, is the parametric Mahalanobis-based score \citep{lee2018mahalanobis}. The Mahalanobis score assumes that the representations from each class are normally distributed around the per-class mean and conform to a mixture of Gaussians \citep{ren2021rmd,fort2021exploring}. Later on, \cite{sun2022oodknn} established an important yet simple OOD detection score, namely the $k$-nearest neighbors (NN) distance, without requiring the feature norms \citep{tack2020csi} or temperature tuning \citep{rafiee2022sdns}. The $k$-NN distance has the advantage of being non-parametric and model- and distribution-agnostic.

Supervised learning may not always produce sufficiently informative features for identifying OOD samples \citep{winkens2020contrastive}. To this end, additional tasks have been proposed to enrich the supervised-learned features. Examples include determining the key in-distribution transformations and predicting them \citep{hendrycks2019rotpred} or contrastive learning. \cite{nvidia2021shifting} attempt to first learn the domain-specific transformations for each in-distribution using Bayesian optimization. \cite{zhang2020hybrid} present a two-branch framework, where a generative flow-based model and a supervised classifier are jointly trained.

\subsection{Unsupervised OOD detection methods}
Unsupervised OOD detection methods rely on learning in-distribution features, typically accomplished with contrastive or supervised contrastive learning \citep{sehwag2021ssd, khosla2020supsimclr}. Contrastive-based methods can be further enhanced by designing hand-crafted transformations that provide an estimate of near OOD data \citep{rafiee2022sdns} or by developing better OOD detection scores \cite{tack2020csi}. For example, \cite{tack2020csi} add a transformation prediction objective (i.e. rotation prediction such as $[0\degree, 90\degree, 180\degree, 270\degree]$). \cite{sehwag2021ssd} define a simpler contrastive-based OOD detection method, where the Mahalanobis distance is computed in the feature space using the cluster centers of $k$-means \citep{lloyd1982kmeans}.

\subsection{OOD detection methods using external data or pretrained models}
Early works attempted to use external data to generate examples near the OOD decision boundary or incorporate them for outlier exposure and negative sampling \citep{hendrycks2018deep, rafiee2022sdns}. Nonetheless, disjointness between the in and out-distribution cannot be guaranteed, and applying shifting transformations depends on the in-distribution \citep{nvidia2021shifting}, which limits the applicability of such approaches.

\cite{hendrycks2020pretrained} show that large-scale models pretrained on diverse external datasets can boost OOD detection performance. Recent OOD detection scores, which leverage pretrained models, deal with the large semantic space by, for instance, grouping images with similar concepts into small groups as in \cite{huang2021mos}. The majority of existing methods focus on supervised OOD detection \citep{sun2022oodknn,galil2023a}. Such approaches include fine-tuning the whole network or parts of it \citep{reiss2021panda}. Contrarily, \cite{ren2021rmd} showed that a Mahalanobis-based score could achieve comparable OOD detection performance on small-scale benchmarks without fine-tuning. 

Out of the limited label-free methods based on pretrained models, a simple approach is the $k$-NN feature similarity, which has not been studied systematically across pretrained models \citep{sun2022oodknn}. Another unsupervised approach aims at initially detecting an \textit{a priori} set of visual clusters. The obtained clusters are subsequently used as pseudo-labels for fine-tuning \citep{cohen2023out}. Apart from unsupervised OOD detection, CLIP models can additionally leverage in-distribution class names \citep{esmaeilpour2022zero}, and in-distribution prototypes can be obtained using the textual encoder, which has not been thoroughly investigated. \cite{esmaeilpour2022zero} extend the CLIP framework by training a text-based generator on top, while \cite{ming2022mcm} compute the maximum softmax probability of image-text feature similarities. Approaches that would alleviate the need to fine-tune billion-scale models remain relatively unexplored, especially in conjunction with CLIP models \cite{wortsman2022robust}.

\subsection{OOD detection robustness}
\cite{hendrycks2018oodrobust} analyzed the robustness under corruptions and geometric perturbations, such as Gaussian noise and brightness shift. Since it is not always clear which manually perturbed images are present in the in-distribution and which are not, attention has been given to adversarial robustness \citep{chen2020robust}. Existing works have primarily focused on fooling supervised OOD detection methods on small-scale benchmarks \citep{adversarialAdam}. Despite the fact that CLIP models' zero-shot classification performance deteriorates significantly when the input images are constructed adversarially \citep{mao2023understanding}, their adversarial OOD detection robustness has not been thoroughly investigated.

\section{The proposed OOD detection setup}
\subsection{Considered pretrained models}
Several supervised CNNs, such as ResNet50 \citep{resnet}, ConvNext \citep{convnext} and ViT \citep{vit} models trained on ImageNet and ImageNet-21K \citep{deng2009imagenet,russakovsky2015imagenet} were utilized. Regarding Imagenet-pretrained self-supervised models, the DINO \citep{dino}, MoCov3 \citep{mocov3}, and MSN \citep{msn} were selected. Finally, CLIP-based models were either trained on OpenAI-400M \citep{radford2021clip} or LAION-2B \citep{schuhmann2022laion}, which consists of 400M and 2 billion image-text description pairs, respectively. Further information regarding the network complexities is reported in \cref{tab:arch_numbers}. To quantify the complexity, we performed inference on a single GPU with a batch size of $256$ to compute the images per second that were processed at $224 \times 224$  resolution.

\begin{table}
\begin{minipage}{.4\textwidth}
\begin{center}
\begin{tabular}{l@{\hspace{0.0cm}}cc}
    \toprule
    \textbf{Architecture} &  \multicolumn{1}{m{16mm}}{\textbf{Images/}} &  \textbf{Number of} \\
     &  \multicolumn{1}{m{16mm}}{\textbf{second}} &  \textbf{Params (M)} \\
    \midrule
    ResNet50        &             2719 &              24  \\
    ConvNext-S      &             1576 &              49  \\
    ConvNext-B      &             1157 &              88  \\
    ConvNext-L      &              695 &             196  \\
    ConvNext-B wide &              888 &              88  \\
    ConvNext-L deep &              531 &             200  \\
    ConvNext-XXL    &              180 &             847 \\
    ViT-S/16        &             3210 &              22  \\
    ViT-B/16        &             1382 &              86  \\
    ViT-L/16        &              475 &             303  \\
    ViT-H/14        &              183 &             631  \\
    ViT-G/14        &               78 &            1843  \\
    \bottomrule
    \end{tabular}
\end{center}
\caption{\textbf{Number of parameters (in millions) and inference time (images per second on a single GPU) for the utilized network architectures.}}
\label{tab:arch_numbers}
\end{minipage}
\hfill %
\begin{minipage}{0.5\textwidth}
\begin{center}
\begin{tabular}{lccc}
\toprule
\textbf{Dataset} & \textbf{Classes} & \textbf{Train} & \textbf{Validation} \\ 
 &  & \textbf{images} & \textbf{images} \\ 
\midrule
\multicolumn{4}{l}{\textit{Pretraining datasets}}\\	
ImageNet & 1K & 1.28M & -\\
ImageNet-21K & 21K &  14M & - \\
OpenAI-400M & - & 400M & - \\
LAION-2B & - & 2B & - \\
\hline
\multicolumn{4}{l}{\textit{In-distribution datasets}}\\	
CIFAR10 & 10 & 50K & 10K \\
CIFAR100 & 100 & 50K & 10K\\
ImageNet & 1K & 1.28M & 50K \\
\hline
\multicolumn{4}{l}{\textit{Out-distribution datasets}}\\
iNaturalist & 110 & - & 10K \\
SUN & 50 & - & 10K\\
Places & 50 & - & 10K \\
IN-O & 200 & - & 2K\\
NINCO & 64 & - & 5.88K \\
Texture & 47 & - & 5.54K \\
\hline  
CIFAR10-A & 10 & - & 1000 \\
CIFAR10-AS & 10 & - & 1000 \\
\bottomrule
\end{tabular}
\end{center}
\caption{\textbf{An overview of the number of classes and the number of samples on the considered datasets.}}
\label{tab:datasets-info}
\end{minipage}
\end{table}

\subsection{Datasets and metrics}
We denote the in-distribution as $\mathcal{D}_{\text{in}}$, and the out-distribution as $\mathcal{D}_{\text{out}}$. The corresponding train and test splits are indicated with a superscript. To design a fair comparison between vision and vision-language models, we define \emph{unsupervised OOD detection} without having access to the set of $\mathcal{D}_{\text{in}}$ label names. Following \cite{huang2021mos}, we use ImageNet as $\mathcal{D}_{\text{in}}$ for large-scale OOD detection benchmarks and CIFAR for small-scale benchmarks. 

For the large-scale benchmarks, we use the following $5$ OOD datasets: Imagenet-O (IN-O) \citep{hendrycks2021natural}, a flower-based subset of iNaturalist \citep{van2018inaturalist}, Texture \citep{cimpoi2014describing}, a subset of the SUN scene database \citep{xiao2010sun}, a subset of Places \citep{zhou2017places}. IN-O contains 2K samples from ImageNet-21K, excluding ImageNet. It is worth noting that Places, Textures, and IN-O have a class overlap of 59.5\%, 25.6\%, and 20.5\% with ImageNet, according to \cite{bitterwolf2023ninco}. Therefore, we additionally use the newly proposed NINCO \cite{bitterwolf2023ninco}, which contains $5879$ samples belonging to $64$ classes with zero class overlap with ImageNet's classes. Dataset information is summarized in \cref{tab:datasets-info}.

To quantify the OOD detection performance, the area under the receiver operating characteristic curve (AUROC) and the false positive rate at 95\% recall (FRP95) are computed between $\mathcal{D}_{\text{out}}^{test}$ test and $\mathcal{D}_{\text{in}}^{test}$. Below, we present the used OOD detection scores, given a pretrained backbone model $g$.

\textbf{1-NN.} For the unsupervised evaluations, we use the maximum of the cosine similarity \citep{sun2022oodknn} between a test image $x'$ and $x_i \in \mathcal{D}_{\text{in}}^{\text{train}}= \{x_1, x_2 \dots, x_{N} \}$ as an OOD score:
\begin{equation}
    s_{\text{NN}}(x') = \max_{i} \operatorname{sim} (g(x'), g(x_i)),
    \label{eq:score-nn}
\end{equation}
where $\operatorname{sim}(\cdot)$ is the cosine similarity and $N$ the number of $\mathcal{D}_{\text{in}}^{\text{train}}$ samples .

\textbf{Mahalanobis distance (MD).} The MD can be either applied directly on the feature space of the pretrained model, $z_i=g(x_i)$, or on the trained linear head, $z_i=h(g(x_i))$. However, MD assumes that the in-distribution labels $y_i \in \{ y_1, \dots, y_N \}$ are available. We denote the class index $c \in \{1, \dots, C\}$, with $C$ being the number of $\mathcal{D}_{\text{in}}$ classes and $N_c$ the number of samples in class $c$. For each class $c$, we fit a Gaussian distribution to the representations $z$ \citep{lee2018mahalanobis}. Specifically, we first compute the per-class mean $\mu_c = \tfrac{1}{N_c} \sum_{i:y_i=c} z_i$ and a shared covariance matrix 
\begin{equation}
\Sigma = \frac{1}{N}\sum_{c=1}^C \sum_{i:y_i=c} (z_i-\mu_c) ( z_i-\mu_c)^{\top}.  
\label{eq:covariance-maha}
\end{equation}

The Mahalanobis score is then computed for each test sample as
\begin{align}
\operatorname{MD}_c(z') &=  \bigl(z'-\mu_c\bigr) \Sigma^{-1} \bigl(z'-\mu_c\bigr)^\top, \label{eq:md-dist} \\
    s_{\text{MD}}(x') &= - \min_c \operatorname{MD}_c(z')\, .
    \label{eq:score-md}
\end{align}

MD can also be applied with cluster-wise means, for instance, using the $\textit{k}$-means cluster centers computed on the feature space of $g$ \citep{sehwag2021ssd}. We denote this score as $\textit{k}$-means MD and use the number of ground truth $\mathcal{D}_{\text{in}}$ classes to compute the cluster centers.

\textbf{Relative Mahalanobis distance (RMD).} Given the in-distribution mean $\mu_0 = \tfrac{1}{N} \sum_{i}^{N} z_i$, we additionally compute $\Sigma_0 = \frac{1}{N} \sum_{i}^{N} (z_i-\mu_0) ( z_i-\mu_0)^{\top}$ to compute  $\operatorname{MD}_0$ analogously to \cref{eq:md-dist}. Subsequently, the RMD score \cite{ren2021rmd} can be defined as
\begin{equation}
    s_{\text{RMD}}(x') = - \min_c \bigl(\operatorname{MD}_c(z')-\operatorname{MD}_0(z') \bigr)\, .
    \label{eq:score-rmd}
\end{equation}

\textbf{Energy.} As in \cite{liu2020energy}, negative free energy is computed over the logits $z$ of an $\mathcal{D}_{\text{in}}$ classifier as  
\begin{equation}
    \label{eq:energy_softmax}
    s_{\text{energy}}(x') = T \cdot \text{log}\sum_i^C e^{z'_i/T},
\end{equation}
where $T=1$ is a temperature hyperparameter.

\subsection{Pseudo-Label Probing (PLP) using CLIP's textual encoder}
\textbf{The pseudo-MSP baseline.} \cite{ming2022mcm} proposed a simple visual OOD detection method to leverage CLIP by feeding the $\mathcal{D}_{\text{in}}$ class names to CLIP's text encoder. The resulting text representations are utilized to compute the cosine similarities for each test image, and then the maximum softmax probability (MSP) is computed as an OOD score. We use pseudo-MSP as a baseline.

\textbf{PLP.} We propose a simple method called \textit{pseudo-label probing} (PLP) that extends \cite{ming2022mcm} by first computing the indices corresponding to the maximum image-text similarity to derive a pseudo-label for each $\mathcal{D}_{\text{in}}^{train}$ image. A linear layer is then trained with the derived pseudo-labels on the features of $g$ and evaluated using the RMD or energy score, namely \emph{PLP + RMD} and \emph{PLP + Energy}. PLP remains OOD data-agnostic and assumes no prior information on $\mathcal{D}_\text{out}$ while training a linear layer adds minimal overhead compared to pseudo-MSP. PLP is conceptually similar to self-training \cite{xie2020self}. Differently from standard self-training, we only train a linear layer using the text-based prototype that is closest to each $\mathcal{D}_{\text{in}}$ image in feature space derived from the textual encoder of CLIP. The obtained text-based prototypes are treated as pseudo-labels, and a linear mapping is learned from the features of $g$ to the $\mathcal{D}_{\text{in}}$ classes.

Similarly, we consider supervised linear probing as an alternative to the existing paradigm of fine-tuning visual feature extractors. Probing refers to training a linear head on the features of the backbone $g$, using the $\mathcal{D}_{\text{in}}^{\text{train}}$ labels and acts as an upper bound for PLP. Subsequently, RMD or Energy is computed. This is in contrast to existing approaches \citep{fort2021exploring,huang2021mos} that typically fine-tune the visual backbone, which is significantly slower, computationally more expensive, and may lead to catastrophic forgetting \citep{kemker2018measuring}.

\begin{figure}
    \centering
    \begin{subfigure}{0.16\columnwidth}
        \includegraphics[width=\textwidth]{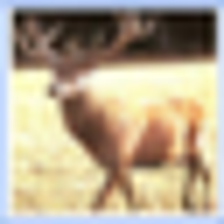}
        \caption{}
    \end{subfigure}%
    \begin{subfigure}{0.16\columnwidth}
        \includegraphics[width=\textwidth]{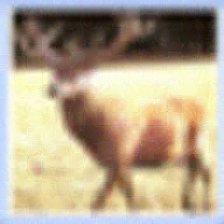}
        \caption{}
    \end{subfigure}%
    \begin{subfigure} {0.16\columnwidth}
        \includegraphics[width=\textwidth]{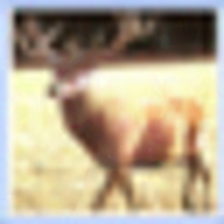}
        \caption{}
    \end{subfigure}
    \begin{subfigure}{0.16\columnwidth}
        \includegraphics[width=\textwidth]{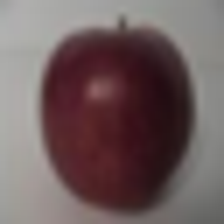}
        \caption{}
    \end{subfigure}%
    \begin{subfigure}{0.16\columnwidth}
        \includegraphics[width=\textwidth]{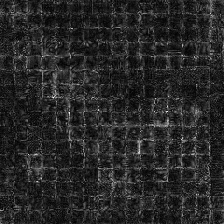}
        \caption{}
    \end{subfigure}%
    \begin{subfigure}{0.16\columnwidth}
        \includegraphics[width=\textwidth]{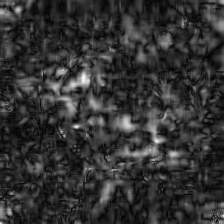}
        \caption{}
    \end{subfigure}
    \caption{\textbf{Generating an adversarial OOD example of a {deer} that is close enough in the feature space to an in-distribution image of an apple}. From left to right: a) the original OOD image from the CIFAR10 test set, b) the adversarial example without smoothing, c) the adversarial example with smoothing, d) the randomly sampled in-distribution target image from CIFAR100, e) the per-pixel Euclidean distance between the original image and perturbed image, and f) the distance between the original and the smoothly perturbed image.}
    \label{fig:adversarial}
\end{figure} 
\subsection{Feature-based adversarial OOD data manipulation}
The unsupervised OOD detection performance of CLIP ViT-G, combined with its known classification robustness against natural distribution shifts, raises the question of whether these models are also adversarially robust OOD detectors. To answer this question, we develop an adversarial OOD data manipulation method that matches image features. In particular, for a test image $x' \in \mathcal{D}_\text{out}^\text{test}$, we randomly choose an in-distribution image $x \in \mathcal{D}_\text{in}^\text{train}$ as the target. In contrast to existing adversarial attacks \citep{adversarial-ood, chen2020robust} that maximize the OOD scores directly, we create an adversarial perturbation $\rho$ with the same dimensions as $x'$ that maximizes the cosine similarity between the in-distribution feature $g(x)$ and $g(x' + \rho)$. Our method targets feature similarities, and it is thus more general as it affects all OOD detection scores. We use the Adam optimizer to compute $\rho$ by minimizing $-\operatorname{sim}(g(x), g(x' + \rho))$, starting with Gaussian noise $\rho \sim \mathcal{N}(0, 10^{-3})$ and clipping $x' + \rho$ to the pixel range $[0, 1]$ after every update step, similar to \cite{adversarialAdam}. We emphasize that we do not directly restrict the perturbation size and only limit the number of steps, as opposed to \cite{adversarialAdam}.

\begin{table*}[t]
\centering
\resizebox{\textwidth}{!}{
\begin{tabular}{lcccccccccc}
\toprule
\multirow{3}{*}{ {Method}} & \multicolumn{9}{c}{OOD Dataset}                        &  \\
                        & \multicolumn{2}{c}{iNaturalist Plants} & \multicolumn{2}{c}{SUN} & \multicolumn{2}{c}{IN-O} & \multicolumn{2}{c}{Texture} & \multicolumn{2}{c}{NINCO}            \\
                        \cmidrule(lr){2-3}\cmidrule(lr){4-5}\cmidrule(lr){6-7}\cmidrule(lr){8-9}\cmidrule(lr){10-11}
                        &  {FPR95$\downarrow$}         &  {AUROC$\uparrow$}      &  {FPR95$\downarrow$}           &  {AUROC$\uparrow$}         &  {FPR95$\downarrow$}          &  {AUROC$\uparrow$}        &  {FPR95$\downarrow$}         &  {AUROC$\uparrow$}      &  {FPR95$\downarrow$}          &  {AUROC$\uparrow$}        \\  
                        \midrule

\multicolumn{11}{c}{\textbf{ CLIP ViT-B}}          \\

\cite{ming2022mcm} & \textbf{30.91}	& \textbf{94.61}	& \textbf{37.59}& 	\textbf{92.57} & 	-& 	-& 	57.77& 	86.11& 	- & 	-\\
Pseudo-MSP$^{\dagger}$  & 52.15 & 89.55 & 40.34 & 91.93 & 78.20 & 79.47 & 55.36 & \textbf{86.82} & 83.09 & 72.44 \\ 
PLP + Energy & 47.70 & 93.36 & 58.15 & 88.69 & \textbf{79.00} & \textbf{80.72} & \textbf{54.63} & {86.44} & \textbf{81.93} & \textbf{75.65} \\
\midrule
\multicolumn{11}{c}{\textbf{ CLIP ViT-L}}          \\
\cite{ming2022mcm} &       28.38&	94.95&	\textbf{29.00}&	\textbf{94.14}&	- &	- &	59.88&	84.88&-&	-\\
Pseudo-MSP$^{\dagger}$ &   48.30 & 90.99 & 29.83 & 93.91 & 72.15 & 81.49 & 57.93 & 85.42 & 74.72 & 78.47 \\
PLP + Energy     & \textbf{27.24} & \textbf{95.40} & 44.99 & 91.05 & \textbf{64.70} & \textbf{85.81} & \textbf{46.75} & \textbf{88.22} & \textbf{71.95} & \textbf{83.83} \\

\midrule \midrule
\multicolumn{11}{c}{\textbf{ CLIP ViT-H (LAION-2B) }} \\   
Pseudo-MSP$^{\dagger}$ & 60.35 & 90.45 & 42.92 & 91.50 & 59.65 & 87.00 & 46.95 & 89.67 & 74.24 & 82.99 \\
 PLP + Energy & \textbf{48.61} & \textbf{93.02} & \textbf{40.90} & \textbf{91.78} & \textbf{52.50} & \textbf{89.24} & \textbf{35.00} & \textbf{91.96} & \textbf{71.9} & \textbf{84.88} \\
\midrule
\multicolumn{11}{c}{\textbf{ CLIP ViT-G (LAION-2B) }} \\   
Pseudo-MSP$^{\dagger}$ & 56.49 & 90.55 & 38.69 & 92.75 & 55.50 & 88.11 & 46.95 & 89.70 & 71.30 & 83.38 \\
 PLP + Energy & \textbf{23.51} & \textbf{96.01} & \textbf{34.41} & \textbf{92.90} & \textbf{42.80} & \textbf{91.39} & \textbf{30.38} & \textbf{92.74} & \textbf{59.32} & \textbf{88.65} \\
\bottomrule
\end{tabular}
}
\caption{\small \textbf{OOD detection performance metrics using CLIP models and label names for ImageNet-1K as in-distribution}. We report the best metric in bold across different ViTs. The symbol $^{\dagger}$ indicates our reproduction of \cite{ming2022mcm}. Crucially, we used the official weights from \cite{radford2021clip} while Ming et al. used the weights provided by Hugging Face trained on different web-crawled data, which justifies the discrepancy between the reported results.}
\label{tab:1k-experiments}

\end{table*}
 
We experimentally observe that in the case of ViTs, the perturbations are quite visible along the edges of the transformer patches ( \cref{fig:adversarial}). To create more natural appearing adversarial examples, we enforce the smoothness of the perturbation by regularizing the allowed perturbation difference between neighboring pixels. We compute the image gradient $\partial \rho / \partial h$ and $\partial \rho / \partial w$ in the horizontal and vertical direction, respectively. The image gradients have the same shape as the image, $3\times H \times W$, and we define the regularization term as
\begin{equation}
    \ell_\text{smooth}(\rho) = \frac{1}{3HW} \sum_{ijk} \left(\frac{\partial \rho}{\partial h}\right)_{ijk}^2
    + \left(\frac{\partial \rho}{\partial w}\right)_{ijk}^2,
\end{equation}
where $i,j,k$ run over image dimensions. We then minimize the loss
\begin{equation}
 \ell_\text{adv} = -\operatorname{sim}(g(x), g(x' + \rho)) + \lambda \ell_\text{smooth}(\rho), 
 \label{eq:adv-smooth}
\end{equation}
with respect to the pertubation $\rho$, where $\lambda$ is a hyperparameter. During the evaluation, we remove the chosen target image, $x$, from $\mathcal{D}_\text{in}^\text{train}$ to show that the adversarial example, $x'+\rho$, cannot be detected as OOD from the remaining in-distribution examples. As a proof of concept, we create two adversarial OOD datasets\footnote{https://drive.google.com/drive/folders/1pYGEPQwagRzdKlPqMQv7sRH6C9MV5vNF} for the CIFAR100 $\rightarrow$ CIFAR10 benchmark, namely CIFAR10-A ($\lambda=0)$ and its smoothened version CIFAR10-AS ($\lambda>0$). The generation of an adversarial example is shown in \cref{fig:adversarial}. More adversarial examples can be found in the supplementary material.

\subsection{Experimental evaluations}
First, we benchmark $25$ publicly available pretrained models on common ImageNet and CIFAR OOD detection benchmarks, as illustrated in Fig. \ref{fig:diagram}. Second, we compare PLP against the aforementioned baselines on ImageNet benchmarks in \cref{tab:1k-experiments}. For a fair comparison with \cite{ming2022mcm}, we use the CLIP ViT-B and ViT-L trained on OpenAI-400M \cite{radford2021clip}, but also report results with the recently released CLIP ViT-G trained on LAION-2B \citep{cherti2022reproducible}. In contrast to \cite{huang2021mos}, we remove the Places OOD subset due to its high class overlap of 59.5\% with ImageNet \citep{bitterwolf2023ninco}; instead, we add IN-O and NINCO. Third, in \cref{tab:1k-supervised}, we compare linear probing on CLIP's visual representations to standard supervised fine-tuning for OOD detection. Fourth, in \cref{tab:ood-evals-vit-g}, we conduct further OOD detection evaluations with CLIP ViT-G, based on the availability of $\mathcal{D}_{\text{in}}$ class names or (few-shot) labeled images and different detection scores. In \cref{tab:cifar-benchmark}, we compare linear probing to fine-tuning for ImageNet-21K pretrained models. Finally, we study the robustness against the adversarially created OOD samples (CIFAR10-A, CIFAR10-AS) using CIFAR100 as $\mathcal{D}_{\text{in}}$.

\subsection{Implementation details}

\label{sec:implementation}
Since probing and PLP only train a linear layer on precomputed representations, it is more scalable and significantly faster than fine-tuning while having minimal overhead compared to pseudo-MSP. We used the Adam optimizer \citep{adam} with a mini-batch size of $256$ for CIFAR10 and CIFAR100 and $8192$ for ImageNet and trained for $100$ epochs with a weight decay of $10^{-3}$. The learning rate is set to $10^{-3} \cdot (\text{mini-batch size})/256$ with a linear warm-up over the first ten epochs and cosine decay after that. All the experiments were carried out in a single NVIDIA A100 with $40$GB VRAM. Moreover, we emphasize that the standard deviation of probing and PLP is less than $0.01$\%, measured over $10$ independent runs. To create the adversarial datasets CIFAR10-A and CIFAR10-AS, we perform $250$ steps with the Adam optimizer with a learning rate of $10^{-3}$ on 1K OOD images. We set $\lambda$ to $5 \cdot 10^{3}$ when applying smoothing (Eq. \ref{eq:adv-smooth}). 

\begin{table*}[t]
\centering
\resizebox{\textwidth}{!}{
\begin{tabular}{lcccccccccc}
\toprule
\multirow{3}{*}{ {Method}} & \multicolumn{9}{c}{OOD Dataset}                        &  \\
                        & \multicolumn{2}{c}{iNaturalist Plants} & \multicolumn{2}{c}{SUN} & \multicolumn{2}{c}{Places} & \multicolumn{2}{c}{Texture} & \multicolumn{2}{c}{NINCO}            \\
                        \cmidrule(lr){2-3}\cmidrule(lr){4-5}\cmidrule(lr){6-7}\cmidrule(lr){8-9}\cmidrule(lr){10-11}
                        &  {FPR95$\downarrow$}         &  {AUROC$\uparrow$}      &  {FPR95$\downarrow$}           &  {AUROC$\uparrow$}         &  {FPR95$\downarrow$}          &  {AUROC$\uparrow$}        &  {FPR95$\downarrow$}         &  {AUROC$\uparrow$}      &  {FPR95$\downarrow$}          &  {AUROC$\uparrow$}        \\  
\midrule
\multicolumn{11}{c}{\textbf{CLIP ViT-B}} \\
Finetune + MSP & 36.80 &90.49 &	60.53&	81.72 &	 63.29 & 80.54  & 54.42 & 82.60 &	\textbf{63.70} & \textbf{80.94} \\
Probing + Energy  & \textbf{24.36} & \textbf{95.72} & \textbf{50.30} & \textbf{90.03} & \textbf{50.26} & \textbf{88.60} & \textbf{50.97} & \textbf{88.10} & 78.31 & 79.23\\
\midrule

\multicolumn{11}{c}{\textbf{CLIP ViT-L}} \\
Finetune + MSP & 31.33 & 91.90 &	51.33& 84.97 &	55.68 & 82.90 &	48.68& 84.52  &	\textbf{56.79} &	84.27 \\ 
Probing + Energy   & \textbf{8.65} & \textbf{97.89} & \textbf{41.42} & \textbf{91.91} & \textbf{43.05} & \textbf{90.81} & \textbf{44.14} & \textbf{90.27} & {67.05} & \textbf{86.32} \\
\midrule  \midrule  

\multicolumn{11}{c}{\textbf{CLIP ViT-H (LAION-2B) }} \\
Finetune + MSP& 26.70 & 	92.90 &	49.19 &	 85.69 &	54.51&	83.28 &	 46.66 & 84.70 &	55.48 &	{83.90} \\ 
Probing + Energy   & \textbf{6.46} & \textbf{98.28} & \textbf{32.42} & \textbf{93.47} & \textbf{38.89} & \textbf{91.68} & \textbf{26.45} & \textbf{93.93} & \textbf{55.39} & \textbf{89.79} \\
\midrule
\multicolumn{11}{c}{\textbf{CLIP ViT-G (LAION-2B) }} \\
Probing + Energy  & 6.29 & 98.27 & 32.21 & 93.27 & 37.16 & 92.02 & 24.63 & 94.31 & 49.35 & 90.4 \\
\bottomrule
\end{tabular}}

\caption{\small \textbf{Supervised OOD detection performance on ImageNet-1K using CLIP: linear probing outperforms fine-tuning on average, especially for larger-scale models, while being computationally cheaper and significantly faster to train}. The best method for each model architecture is bolded, excluding CLIP ViT-G. The reported results from fine-tuning are based on our evaluations using publicly available CLIP weights fine-tuned on ImageNet-1K from \textit{timm} \cite{rw2019timm}.}
\label{tab:1k-supervised}

\end{table*}

\section{Experimental results}
\textbf{Unsupervised OOD detection.} In \cref{fig:diagram}, we initially investigate whether there is a connection between the $\mathcal{D}_{\text{in}}^{test}$ classification accuracy and unsupervised OOD detection AUROC by benchmarking $25$ feature extractors. Out of them, CLIP models exhibit the strongest correlation ($R^2$ coefficient $\geq0.92$) independent of their network architecture (i.e. ConvNext, ViTs, etc). CLIP's best instances (ViT-G, ConvNext-XXL) are currently the best-performing unsupervised OOD detectors, aligning with recent results from \cite{galil2023a}. The features of CLIP ViT-G even outperform the ones from supervised training on ImageNet as $\mathcal{D}_{\text{in}}$. Moreover, we observe that when the $\mathcal{D}_{\text{pretrain}}$ is different from $\mathcal{D}_{\text{in}}$ (CIFAR benchmarks), a positive correlation can still be identified for self-supervised and supervised models pretrained on ImageNet ($R^2 = 0.82$), yet not as strong as CLIP ($R^2 = 0.95$). 

\textbf{Large-scale OOD detection with access to class names.} We report absolute gains and AUROC scores. In \cref{tab:1k-experiments}, we compare pseudo-MSP, based on our reproduction of \citet{ming2022mcm}, to PLP on all five large-scale benchmarks and find an average improvement of 2.81\%, 1.85\%, 3.44\% AUROC for CLIP ViT-L, ViT-H, and ViT-G, respectively. The reported gains are computed using the energy score as it was slightly superior to RMD on average and significantly faster to compute. Interestingly, PLP has consistent improvements on the newly proposed NINCO dataset, wherein all samples have been visually verified to be semantically different from $\mathcal{D}_{\text{in}}$, with an average AUROC gain of 3.93\% across CLIP models. We highlight that the discrepancy between our reproduction and the reported results of \citet{ming2022mcm} using ViT-B and ViT-L is due to different CLIP weights: the authors used weights provided by Hugging Face trained on different web-crawled data. In contrast, we used the official weights from \cite{radford2021clip}.  

\textbf{Large-scale supervised OOD detection.} As shown in  \cref{tab:1k-supervised}, fine-tuning CLIP models is often unnecessary for supervised OOD detection. More precisely, probing on larger models such as CLIP ViT-H consistently outperforms fine-tuning by a large margin of 7.34\% on average. The performance discrepancy of CLIP ViT-L on Places and SUN compared to the other benchmarks is partially attributed to the class overlap. By contrast, iNaturalist, Texture, and NINCO are sufficiently disjoint from ImageNet in terms of $\mathcal{D}_{\text{in}}$ class overlap. 

\textbf{Adversarial OOD detection robustness.} By evaluating CLIP ViT-G on the introduced CIFAR100$\rightarrow$CIFAR10-A benchmark, we found that it is possible to drop the AUROC score from 87.6\% $\rightarrow$50.3\% using $1$-NN and 94.2\%$\rightarrow$49.5\% using pseudo-MSP. Introducing the smoothness restriction (\cref{eq:adv-smooth}) degrades performance to 55.8\% and 51.9\% AUROC on CIFAR100$\rightarrow$CIFAR10-AS using $1$-NN and pseudo-MSP, respectively. Note that an AUROC score of 50\% is a random guess's score, meaning that CLIP ViT-G performs slightly better than a random guess. 

\begin{table*}
\centering

\begin{tabular}{lccccc}
\toprule
\multirow{ 2}{*}{{CLIP ViT-G/14}} & $\mathcal{D}_{\text{in}}$   
&  $\mathcal{D}_{\text{in}}$:CIFAR100   & CIFAR10 & ImageNet  & ImageNet \\  
& {labels/names}  & 
$\mathcal{D}_{\text{out}}$:CIFAR10 &  CIFAR100 & IN-O & NINCO    \\
\midrule
 $k$-means MD  & \xmark / \xmark & 72.8  &  89.5 & 87.6 &  80.7 \\
$1$-NN                & \xmark / \xmark &  \textbf{87.6} & \textbf{98.2} & \textbf{88.0} & \textbf{84.0} \\
\hline
Pseudo-MSP     & \xmark / \checkmark & 94.2 & 97.3 & 88.1 & 83.4 \\
PLP + MSP & \xmark / \checkmark & 92.7 & 97.9 & 86.6 & 86.7 \\

PLP + RMD & \xmark / \checkmark & \textbf{97.1} & 98.3 & \textbf{91.9} & 88.4 \\

PLP + Energy & \xmark / \checkmark & 95.0 & \textbf{98.5} & 91.4 & \textbf{88.7} \\
\hline

MD  &   \checkmark  /  \checkmark  &   73.1 &  91.1 &  88.1  & 81.5 \\
RMD &   \checkmark  /  \checkmark  &  96.3 &  98.8 &  92.4 & 89.3  \\

Few-shot $p=10$ + MSP &  \checkmark / \checkmark & 89.4 & 96.5 & 88.1 & 84.8 \\
Few-shot $p=10$ + Energy &  \checkmark / \checkmark & 90.9 & 96.6 & 90.8 & 83.0 \\
Probing + MSP  &   \checkmark  / \checkmark  & 94.1 & 98.7 & 88.1 & 89.1 \\
Probing + RMD & \checkmark / \checkmark & \textbf{97.3} & 98.8 & 92.5 & 89.5 \\
Probing + Energy & \checkmark / \checkmark & 96.3 & \textbf{99.1} & \textbf{92.9} & \textbf{90.4} \\

\bottomrule
\end{tabular}
\caption{\textbf{OOD detection AUROCs (\%) for multiple evaluations and scores using CLIP ViT-G/14 trained on LAION-2B.}} 
\label{tab:ood-evals-vit-g}
\end{table*}   

\textbf{Ablation study for CLIP ViT-G for all OOD detection setups.} In \cref{tab:ood-evals-vit-g}, we conduct additional experimental evaluations using CLIP ViT-G for all three OOD detection scenarios (unsupervised, class names are available, supervised). In the unsupervised case, $k$-means + MD yields an inferior AUROC compared to $1$-NN, precisely lower by 6.8\% on average. By incorporating the $\mathcal{D}_{\text{in}}$ class names using CLIP’s text encoder, we find that the PLP consistently outperforms pseudo-MSP with a mean improvement of 3.16\% and 2.63\% over PLP+RMD and PLP+Energy respectively. In the supervised scenario, we highlight that RMD is a strong baseline, surpassing MD by 10.75\% AUROC on average, while our PLP+Energy marginally improves RMD by 0.47\%. MSP is constantly the worst choice as an OOD detection score after linear probing.

\section{Discussion}
\textbf{Pixel-related features in CLIP's learned representations.} Similar to \cite{mao2023understanding} for image classification, the OOD detection performance close to a random guess demonstrates that even the top-performing CLIP models trained on billion-scale image-text pairs can be fooled by a visual signal manipulation that is invisible to humans. This finding gives further evidence that besides label-related features, (local) pixel information affects the learned representations and needs to be explored in greater depth \citep{park2023what,dravid2023rosetta}. Another possible research avenue is how adversarial OOD samples transfer between different feature extractors, which is left for future work.

\begin{table*}[t]
\centering
\begin{tabular}{lccccc}
\toprule
\multirow{ 2}{*}{{}}   & Finetuned & $\mathcal{D}_{\text{in}}$:IN1K   & IN1K & CIFAR100 & CIFAR10     \\ 
 &   on $\mathcal{D}_{\text{in}}$  &
    $\mathcal{D}_{\text{out}}$:NINCO &  IN-O  & CIFAR10 & CIFAR100     \\ 
\midrule
ConvNext-B &  \xmark & \textbf{95.4} &  \textbf{94.9} & \textbf{93.0} & \textbf{98.0} \\
ConvNext-B  &   \checkmark & 88.8   &   85.8 & 91.3* & 96.3* \\ 
\hline

ViT-L &  \xmark & \textbf{95.2} & \textbf{95.3} & 89.8 & 94.3 \\
ViT-L &     \checkmark  & 91.3 & 92.1 & \textbf{97.9}* & \textbf{98.5}*\\ 
\hline
ResNet50+ViT-B & \xmark & \textbf{95.9} & \textbf{95.8}  & 89.2 & 96.6 \\
ResNet50+ViT-B & \checkmark & 92.9    &    92.1 &  \textbf{96.2}* & \textbf{98.5}* \\

\bottomrule
\end{tabular}
\caption{\textbf{Comparing the OOD detection AUROC (\%) of ImageNet-21K pretrained models: linear probing versus fine-tuning.} We use publicly available models with and without fine-tuning on ImageNet from \textit{timm} and evaluate them using the energy score. Following \cite{fort2021exploring}, we use the MD on CIFAR fine-tuned models (indicated with an asterisk), which is inefficient at the scale of ImageNet.}
\label{tab:cifar-benchmark}
\end{table*}

\textbf{Does PLP improve $\mathcal{D}_{\text{in}}^{\text{test}}$ accuracy?} By comparing the $\mathcal{D}_{\text{in}}^{\text{test}}$ accuracy on ImageNet of the trained head with the pseudo-labels versus the zero-shot classification of CLIP ViT-G, we found a minor accuracy improvement of $0.5\%$ on ImageNet. However, accuracies deteriorate for CLIP ViT-B and CLIP ViT-L on average, which suggests that their $\mathcal{D}_{\text{in}}$  text-based pseudo-labels have less true positives. The fact that PLP improves the OOD detection performance (compared to pseudo-MSP) without significantly increasing $\mathcal{D}_{\text{in}}^{\text{test}}$ accuracy is a new direction left for future work. Another interesting avenue is whether the text-based pseudo-labels from CLIP can be used with other, possibly smaller-scale, visual feature extractors.  

\textbf{Does linear probing lead to similar OOD detection performance compared to fine-tuning for ImageNet-21K pretrained models?} In \cref{tab:cifar-benchmark}, we compare linear probing versus fine-tuning for $3$ publicly available models, where there is a corresponding fine-tuned model on ImageNet. Even though fine-tuning is the standard practice in supervised OOD detection \cite{fort2021exploring}, the obtained results indicate that it performs inferior to linear probing on the ImageNet OOD-related benchmarks even for ImageNet-21K pretrained models. Fine-tuning is, however, still the best approach on the small-scale OOD benchmarks (e.g. CIFAR10 $\rightarrow$ CIFAR100), given ImageNet21K pretraining. This explains why a simple and scalable approach such as probing may have been overlooked. As suggested by \cite{huang2021mos}, we confirm that OOD detection approaches tested on small-scale benchmarks do not always translate effectively into larger-scale setups, where simpler methods need to be revisited. Combined with the results from \cref{tab:1k-supervised}, we validate that the OOD-related information is readily available on large-scale foundational models, as recently stated in \cite{inkawhich2023adversarial, oquab2023dinov2}.

\begin{figure*}
\begin{center}
\includegraphics[width=0.5\columnwidth]{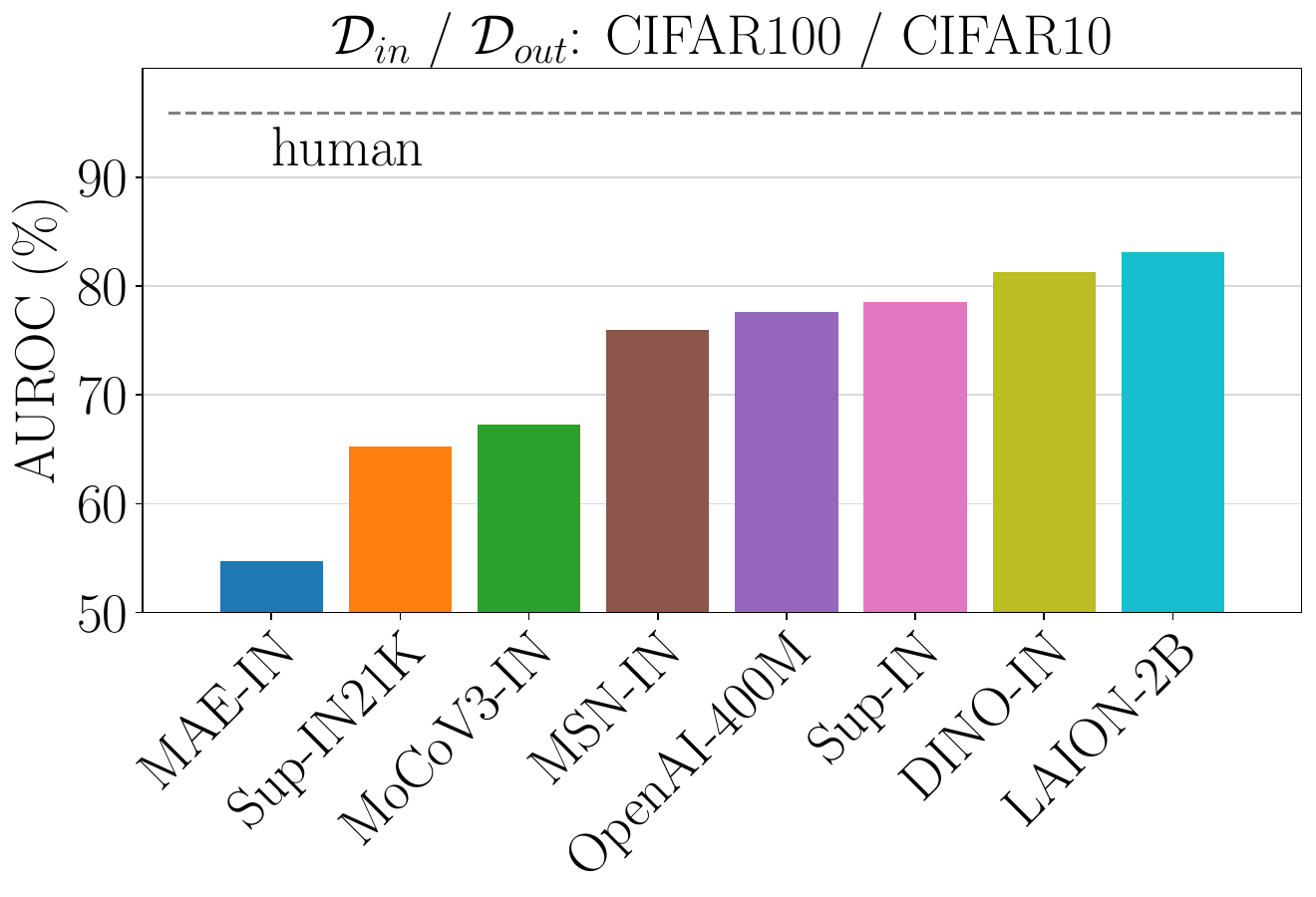}%
\includegraphics[width=0.5\columnwidth]{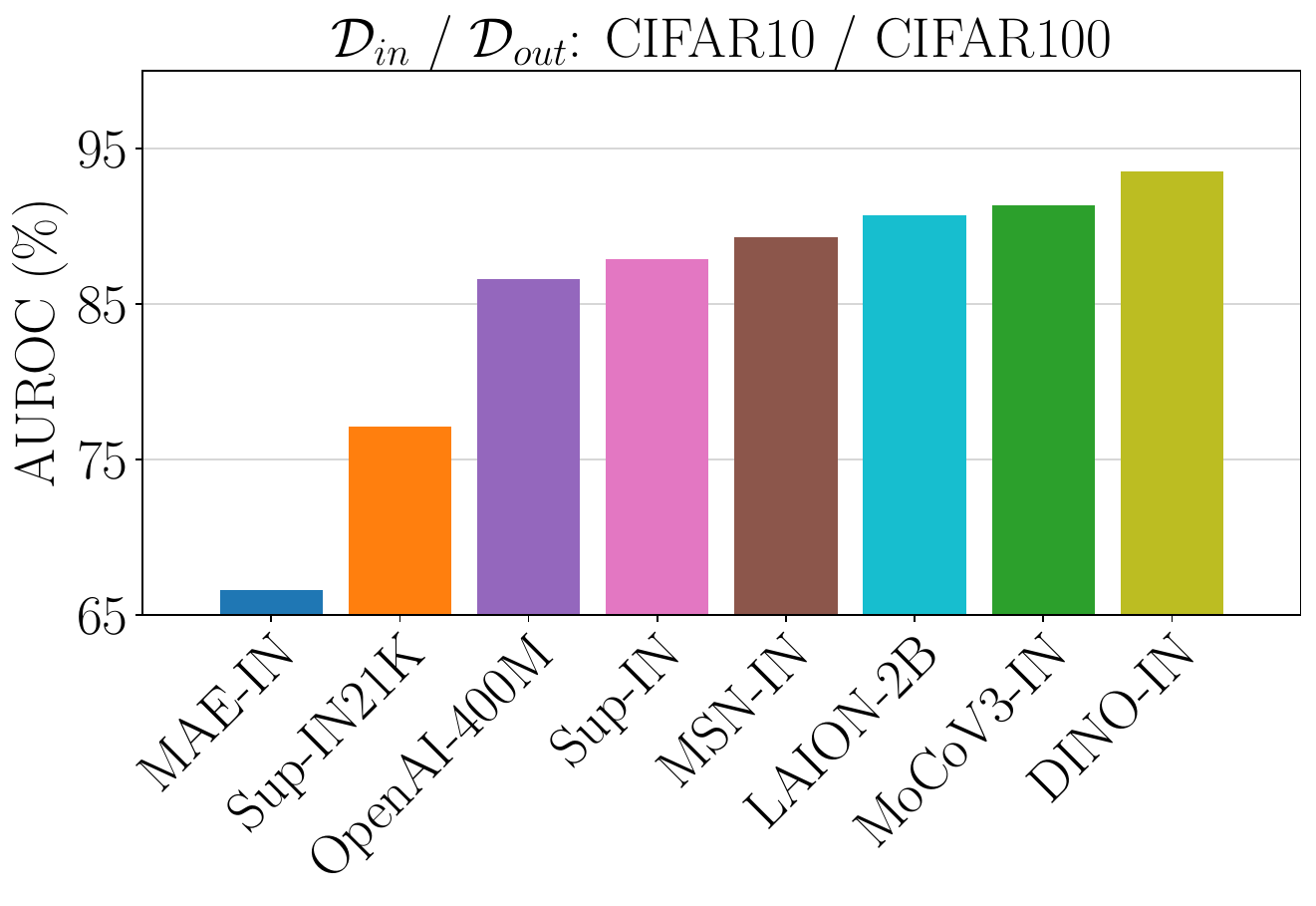}
\end{center}
\caption{\textbf{AUROC values for unsupervised OOD detection using ViT-B/16 pretrained on different datasets (IN, IN-21K, OpenAI-400M, LAION-2B) and pretext tasks}. IN indicates ImageNet. The performance of CLIP on LAION-2B is bottlenecked by the model size as reported in \cite{cherti2022reproducible} and larger networks are needed when scaling up to billion-scale datasets. The horizontal line indicates human-level performance, as reported in \cite{fort2021exploring}.}
\label{fig:pretrain}
\end{figure*}

\textbf{Is the choice of the best feature extractor consistent for smaller-sized models?} Not in the scale of ViT-B and CIFAR benchmarks. In \cref{fig:pretrain}, we keep the model architecture fixed (ViT-B) and visualize the unsupervised OOD detection performance ($1$-NN) for different datasets (ImageNet, ImageNet21K, OpenAI-400M, LAION-2B) and pretraining types (CLIP, supervised, self-supervised). The rationale behind evaluating on the CIFAR-based benchmarks is to focus on feature transferability. We highlight that large pretraining datasets such as LAION-2B are bottlenecked by the model size of ViT-B as explained in \cite{cherti2022reproducible}. Unlike \cref{fig:diagram}, we show that the ranking of backbones is not always consistent between different $\mathcal{D}_{\text{in}}$ and $\mathcal{D}_{\text{out}}$ at this scale. Finally, DINO ViT-B surpasses supervised ImageNet pretraining on both benchmarks, outlining the transferability of features of self-supervised methods, which is consistent with the results of \cite{ericsson2021ssltranfer, naseer2021intriguing, adaloglou2023exploring}.

\section{Conclusion}
This work presented a thorough experimental study by leveraging pretrained models for visual OOD detection, focusing on CLIP. It was demonstrated that CLIP models are powerful unsupervised OOD detectors, outperforming even $\mathcal{D}_{\text{in}}$ supervised models on large-scale OOD detection benchmarks. A fast, simple, OOD data-agnostic, and scalable method called PLP that trains a linear layer based on $\mathcal{D}_{\text{in}}$ text-based pseudo-labels was introduced. PLP outperformed the previous state-of-the-art (pseudo-MSP) for most CLIP models and ImageNet-based benchmarks while substantially improving the larger architectures (i.e. ViT-G achieves a mean AUROC gain of 3.4\%). Furthermore, it was demonstrated that probing can replace the costly fine-tuning on CLIP and even ImageNet-21K models on large-scale benchmarks, where CLIP ViT-H exhibited an average AUROC improvement of 7.3\%. Finally, a novel feature-based adversarial OOD data manipulation method was introduced, which pointed to the fact that billion-scale feature extractors (CLIP ViT-G) are vulnerable to adversarial OOD attacks.

\FloatBarrier
\bibliography{ref}
\bibliographystyle{tmlr}
\clearpage
\FloatBarrier

\appendix

\section{Supplementary material: Adapting Contrastive Language-Image Pretrained (CLIP) Models for Out-of-Distribution Detection}

\subsection{Additional implementation details}
Following \cite{ming2022mcm}, for the text encoder of CLIP, we take the mean of the L2-normalized text representations of the following prompts: ``an image of a \{label\}", ``a photo of a \{label\}", ``a blurry photo of a \{label\}", ``a photo of many \{label\}", ``a photo of the large \{label\}", ``a photo of the small \{label\}". We randomly select $p=10$ images per class for few-shot probing and take the average AUROC over $5$ runs. We set the mini-batch size to $32$ for CIFAR100 and $1024$ for ImageNet. To enforce reproducibility for our results on Table 6, the corresponding models can be found using the \textit{timm} (version 0.6.12) names \cite{rw2019timm}: \textit{vit\_large\_patch16\_224\_in21k}, \textit{convnext\_base\_in22k}, and \textit{vit\_base\_r50\_s16\_224\_in21k}. We note that probing takes less than 15 minutes on ImageNet on a single GPU.

\subsection{Does PLP using an MLP achieve superior results?} 
We found negligible differences when substituting the linear layer with an MLP, which suggests that a linear mapping to the space of in-distribution classes is sufficient for OOD detection.

\subsection{Computational complexity of OOD detection methods.} As reported in Table 6, \cite{fort2021exploring} used the MD as an OOD score, which is inefficient at the scale of ImageNet and can become prohibitively slow for even larger datasets. Given a feature dimension $d$ and dataset size $N$, the total time complexity scales linearly with dataset size (for the computation of the covariance matrix) and in cubic time with $d$ due to the inverse calculation of the covariance matrix, resulting in $O(N d \min(N,d) + d^3)=O(N d^2 + d^3)$ since $N>d$. We thus conclude that, in addition to the performance metrics, the computational complexities need to be taken into account in future studies to design scalable and efficient OOD detection systems.

\subsection{Additional few-shot evaluations on CIFAR100$\rightarrow$ CIFAR10}

\begin{figure}[h]
    \centering
    \includegraphics[width=0.5\columnwidth]{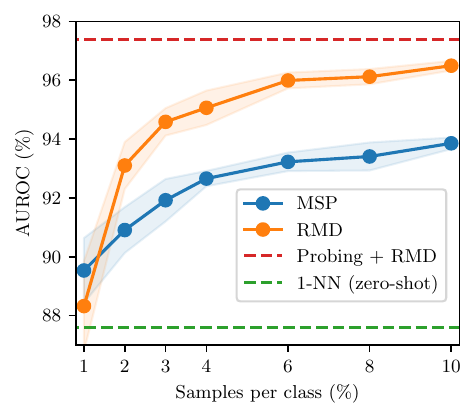}
    \caption{\textbf{Few-shot linear probing on CIFAR100 $\rightarrow$ CIFAR10.} Samples per class (shown as a percentage \%) versus OOD detection performance (\textit{y-axis}).}
    \label{fig:few_shot}
\end{figure}

In Fig. \ref{fig:few_shot}, we show that even with $1$\% of the data, we can surpass the zero-shot $1$-NN score. RMD applied on the logits seems to be consistently better than MSP.

\subsection{Additional adversarial examples}
\begin{figure*}[t]
\includegraphics[width=\textwidth]{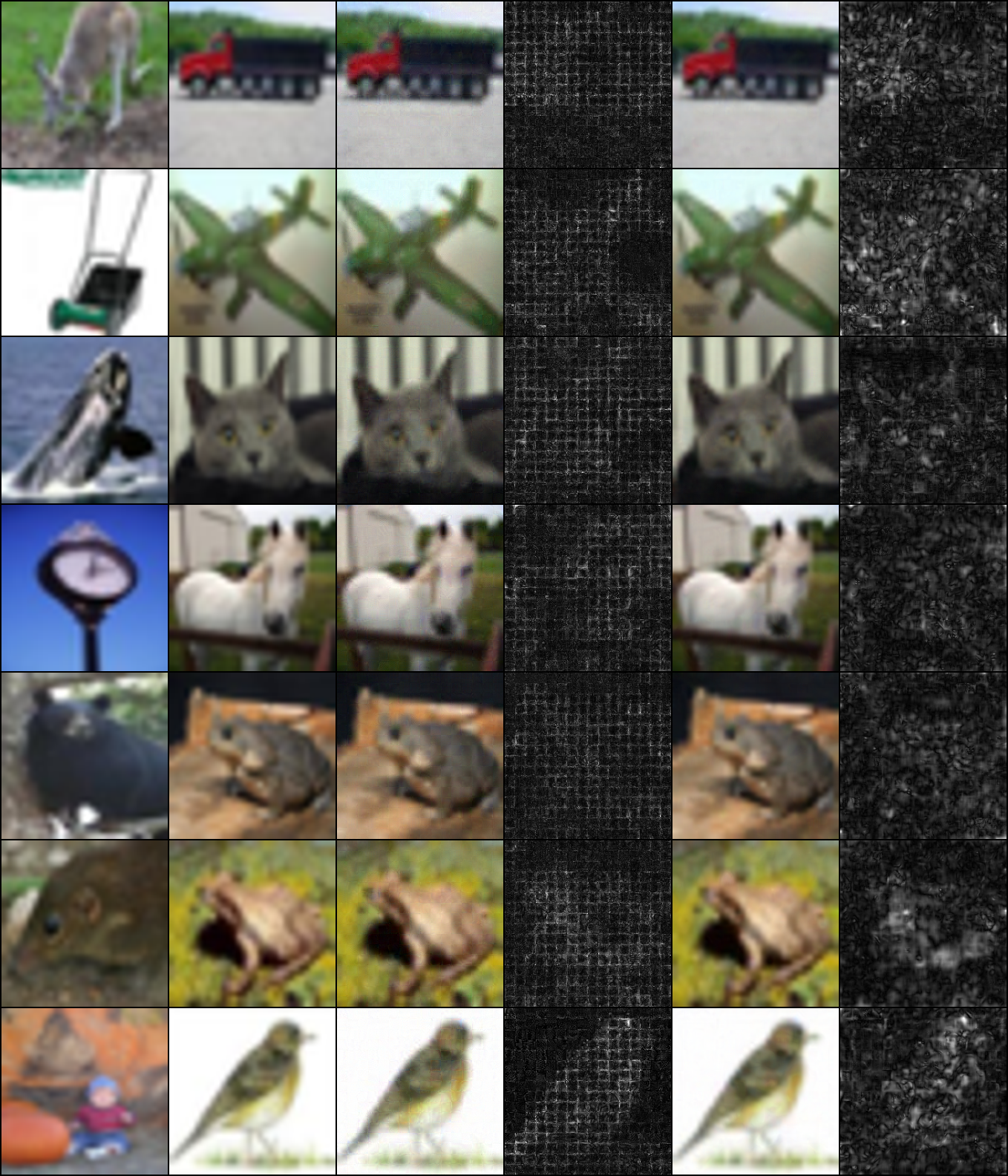}
\caption{\textbf{Additional CIFAR10 adversarially manipulated examples for CIFAR100 $\rightarrow$ CIFAR10 OOD detection with (CIFAR10-AS) and without (CIFAR10-A) the smoothing constraint.} Columns from \textit{left to right}: target in-distribution image from CIFAR100, original CIFAR10 sample, adversarially manipulated image without smoothing, the Euclidean pixel-wise distance between the original image and perturbed image, adversarial example with smoothing, Euclidean distance.}
\label{fig:extra-adversarial-samples}
\end{figure*}

We illustrate more adversarially generated samples using the proposed method in Fig. \ref{fig:extra-adversarial-samples}. The created adversarial OOD datasets CIFAR10-A and CIFAR10-AS are publicly \hyperlink{https://drive.google.com/drive/folders/1pYGEPQwagRzdKlPqMQv7sRH6C9MV5vNF}{available via this hyperlink.}

\end{document}